\documentclass[letterpaper, 10 pt, conference]{ieeeconf}  

\IEEEoverridecommandlockouts                              

\newcommand\copyrighttext{%
	\footnotesize \textcopyright 2020 IEEE.  Personal use of this material is permitted.  Permission from IEEE must be obtained for all other uses, in any current or future media, including reprinting/republishing this material for advertising or promotional purposes, creating new collective works, for resale or redistribution to servers or lists, or reuse of any copyrighted component of this work in other works.\\
	DOI: 10.1109/IROS40897.2019.8967571}

\newcommand\copyrightnotice{%
	\begin{tikzpicture}[remember picture,overlay]
	\node[anchor=north,yshift=-10pt] at (current page.north) {\fbox{\parbox{\dimexpr\textwidth-\fboxsep-\fboxrule\relax}{\copyrighttext}}};
	\end{tikzpicture}%
}

\makeatletter
\@namedef{ver@lastpage.sty}
\@namedef{opt@lastpage.sty}
\makeatother

\usepackage{amsmath}
\usepackage{mathtools}
\usepackage{amsfonts}
\usepackage{graphicx}
\usepackage{bbm}
\usepackage{dsfont}
\usepackage{textcomp}

\usepackage{epstopdf}
\usepackage{units}
\usepackage{pgfplots}
\usepackage{tikz}
\usepgfplotslibrary{external} 
\pgfplotsset{compat=1.14,every tick label/.append style={font=\footnotesize}}
\tikzset{every picture/.style={font issue=\footnotesize},
	font issue/.style={execute at begin picture={#1\selectfont}}
}
\usetikzlibrary{external,arrows.meta,patterns,ipe} 

\newcommand{\partder}[2]{\frac{\partial #1}{\partial #2}}

\DeclareMathOperator*{\argmax}{arg\,max}

\DeclareMathOperator*{\minus}{\!-\!}

\usepackage[caption=false, font=footnotesize]{subfig}

\title{\LARGE \bf
Transfer learning for vision-based tactile sensing}

\author{Carmelo Sferrazza and Raffaello D'Andrea$^{1}$
\thanks{$^{1}$The authors are members of the Institute for Dynamic Systems and Control, ETH Zurich, Switzerland. Email correspondence to Carmelo Sferrazza 
        {\tt\small csferrazza@ethz.ch}}%
}

\begin{document}


\maketitle
\thispagestyle{empty}
\pagestyle{empty}

\copyrightnotice

\begin{abstract}

Due to the complexity of modeling the elastic properties of materials, the use of machine learning algorithms is continuously increasing for tactile sensing applications. Recent advances in deep neural networks applied to computer vision make vision-based tactile sensors very appealing for their high-resolution and low cost. A soft optical tactile sensor that is scalable to large surfaces with arbitrary shape is discussed in this paper. A supervised learning algorithm trains a model that is able to reconstruct the normal force distribution on the sensor's surface, purely from the images recorded by an internal camera. In order to reduce the training times and the need for large datasets, a calibration procedure is proposed to transfer the acquired knowledge across multiple sensors while maintaining satisfactory performance.

\end{abstract}


\section{INTRODUCTION}
The importance of the sense of touch in humans has been repeatedly shown to be fundamental even when all other sensing modalities are available, see for example \cite{touch_importance1} and \cite{touch_importance2}. Similarly, artificial tactile sensors can be of great help to robots for manipulation tasks, and in general for their interaction with the environment and with humans.

The aim of robotic tactile sensing systems is to make a robot capable of sensing the different types of forces, temperature changes and vibrations that its surface is subject to. However, compared to the human tactile sensing system, the state-of-the-art tactile sensors often address only a fraction of these capabilities \cite{tactile_sensing_survey}, and are generally only tailored to specific tasks.

In the last decade, as a result of the increased use of machine learning, computer vision has progressed dramatically. This has enabled the use of cameras to sense the force distribution on soft surfaces, by means of the deformation that elastic materials undergo when subject to force. In this respect, acquisition devices that sense a variety of properties (i.e. force, vibrations) via direct physical contact, using a vision sensor to infer these properties from the change in light intensity or refractive index, are often referred to as optical tactile sensors. In particular, one class of optical tactile sensors relies on tracking the motion of markers distributed over a deformable surface to reconstruct the force distribution, see for example \cite{wenzhen_sensors} and the references therein.

The approach discussed in this article is based on the sensor presented in \cite{sferrazza_sensors}. The sensor exploits the information provided by the movement of spherical markers that are randomly distributed within the volume of a three-dimensional silicone gel. An off-the-shelf camera is used to track the pattern created by these markers inside the gel. In this way, an approximation of the strain field is obtained. Training data is collected with an automatic procedure, where the spherical tip of a needle is pressed against the gel at different positions and depths. During this procedure, the camera is used to collect images of the resulting pattern, while the ground truth normal force is measured with a force torque (F/T) sensor.

\begin{figure}
	\centering
	\includegraphics[width=0.95\linewidth]{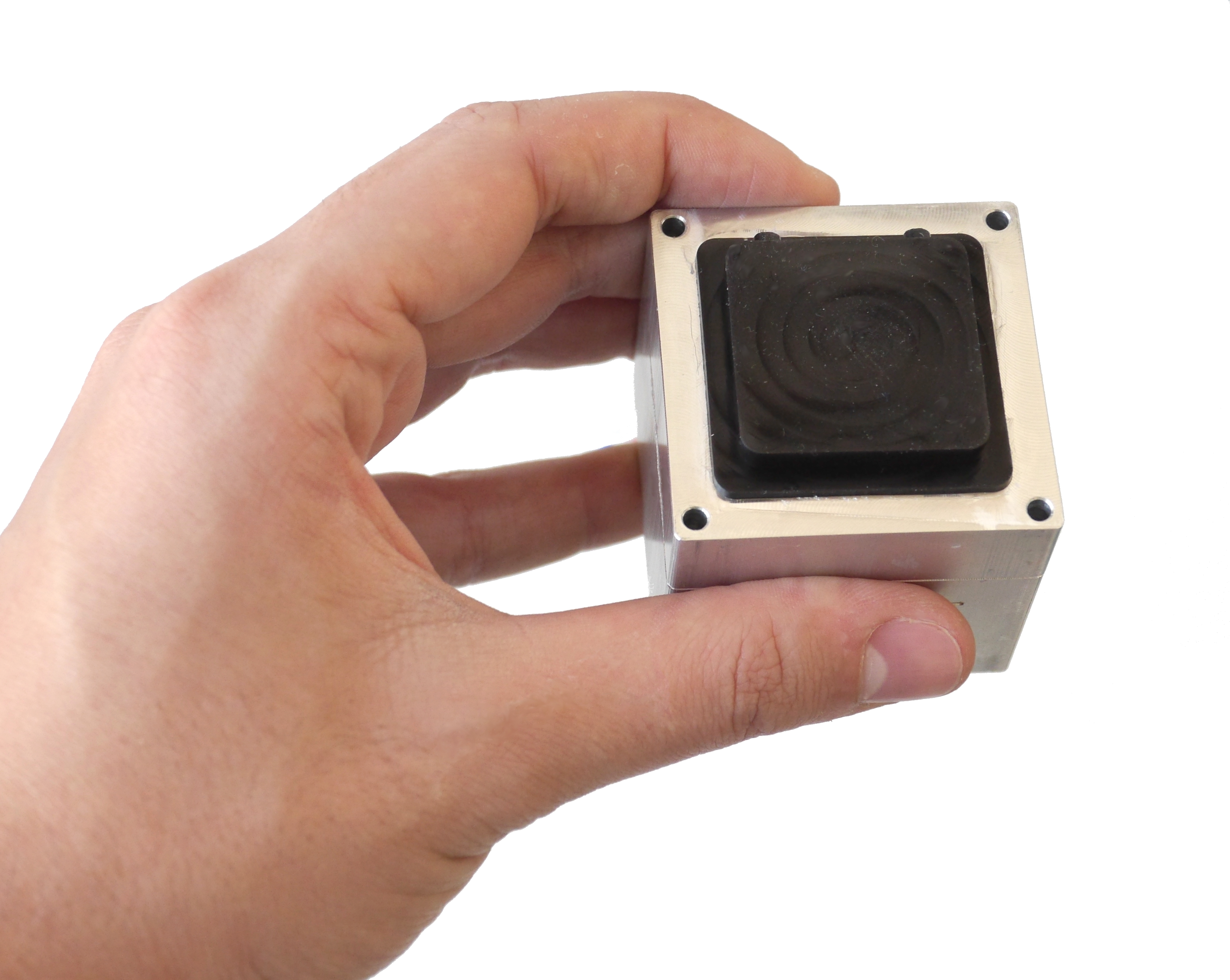}
	\caption{The experimental setup used for the evaluation presented in this article.}
	\label{fig:sensor}
\end{figure}

A neural network is trained with the labeled data to reconstruct the normal force distribution over the deformable sensor's surface. In order to reduce the training data necessary to achieve satisfactory performance on a different gel (with different hardness and marker distribution), a calibration layer is added to the existing architecture, and trained on a considerably smaller dataset. The resulting architecture achieves an accuracy comparable to the one attained by the original model on the larger dataset.

\subsection{Related work}
Sensing the location and the type of the different forces that a body is subject to is a key requirement to enable safe human-robot interaction. However, tactile sensing research is still far from achieving the same sort of comprehensive solution that cameras, with their high resolution and relatively small size and low cost, represent for visual sensing. Various categories of tactile sensors have been proposed in the literature, monitoring and exploiting different properties of materials that change under contact with another body, of which an overview is given in \cite{tactile_sensing_survey} and \cite{tactile_sensing_survey2}.

Compared to other categories (e.g. tactile sensing arrays exploiting changes in capacitance or resistance), soft optical (or vision-based) tactile sensors combine low cost, ease of manufacture and minimal wiring. They intrinsically provide high spatial resolution (stemming from the high resolution of modern image sensors) and preserve the softness of their surface. These advantages come with the drawback of a larger size, which can be overcome by the use of multiple cameras (with a smaller footprint), as conceptually described in \cite{fingervision}.

Vision-based tactile sensors generally observe the behavior of certain light quantities to infer the forces applied to the elastic surface of the sensor. As an example, in \cite{total_internal_reflection}, a tactile sensor that uses the principle of total internal reflection is proposed, while in \cite{gelsight_first} photometric stereo is used to reconstruct the contact shape with a piece of elastometer through the use of differently colored lights.

In \cite{gelforce}, a camera tracks the motion of spherical markers positioned in a grid pattern within a transparent elastic material. An analytical model is used to reconstruct the applied force distribution by approximating the problem with a semi-infinite elastic half-space. In particular, it is experimentally measured how the spread of the markers over two layers at different depths improves the robustness to errors in the computation of the displacement of these markers. A biologically-inspired design is presented and analyzed in \cite{tactip_first} and \cite{tactip_force}, with markers arranged in a special pattern below the surface. The average displacement of these markers is related to the shear and normal forces applied to the contact surface.

However, the complexity of modeling the elastic properties of soft materials makes the derivation of a map from the monitored properties (i.e. marker displacement, changes in light intensity) to force distribution very challenging, especially when a sensor is interfaced with different types of objects. For this reason, machine learning algorithms have recently been applied to the force reconstruction problem, see for instance \cite{jan_peters} and \cite{helge_ritter}.

In \cite{wenzhen_sensors}, a deep neural network is used to estimate the contact force on different types of objects, by using the same sensor as in \cite{gelsight_first} with the addition of a printed layer of markers on the surface.
In \cite{cheeta_mit}, based on an automatic data collection procedure, a neural network estimates the force on a footpad while moving over certain trajectories. 

Compared to most of the approaches cited, the method presented in \cite{sferrazza_sensors} and discussed here does not rely on a special pattern of markers to be tracked by the camera, due to the particular choice of the features extracted from the images. In fact, the random distribution of the markers simplifies manufacture and makes this sensor suitable to adapt to arbitrary shapes and scalable to large surfaces through the use of multiple cameras. The chosen representation of the normal force distribution can handle different types of indenters and multiple points of contact. Moreover, it can easily be extended to sense shear forces. The algorithms proposed here provide an estimate of the normal force distribution at 60 Hz on a standard laptop (dual-core CPU, 2.80 GHz).

Transfer learning, see \cite{transfer_learning_survey} for a survey, is an important topic to address when it becomes relevant to speed up learning. As an example, a data alignment mechanism is proposed in \cite{transfer_learning_peters} to transfer the task knowledge between two different robot architectures.
In the context of learning-based tactile sensing, collecting a dataset might be time consuming. The proposed design and modeling enable the transfer of the knowledge acquired on one tactile sensor across different gels with different marker patterns. To this purpose, a calibration procedure is presented in this paper, which considerably reduces the amount of data required for learning, as well as the training time.

\subsection{Outline}
The sensor design is discussed in Section \ref{sec:sensor_design}. The training data collection procedure is described in Section \ref{sec:data_collection} and the postprocessing of the data into features for the learning algorithm is presented in Section \ref{sec:feature_engineering}. In Section \ref{sec:neural_network}, the neural network architecture is explained and experimental results are presented. The calibration procedure is described in Section \ref{sec:calibration}, while conclusions are drawn in Section \ref{sec:conclusion}.
\section{SENSOR DESIGN} \label{sec:sensor_design}
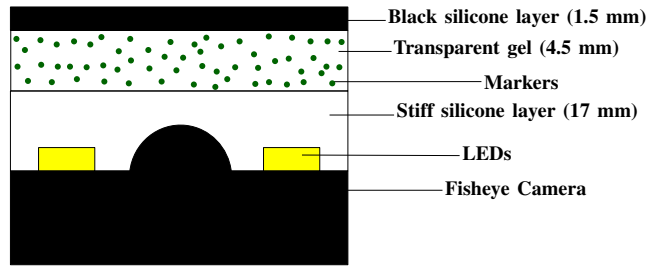
\begin{figure}
	\scalebox{0.38}{\hspace{-3.5cm}\tikzstyle{ipe stylesheet} = [
  ipe import,
  even odd rule,
  line join=round,
  line cap=butt,
  ipe pen normal/.style={line width=0.4},
  ipe pen heavier/.style={line width=0.8},
  ipe pen fat/.style={line width=1.2},
  ipe pen ultrafat/.style={line width=2},
  ipe pen normal,
  ipe mark normal/.style={ipe mark scale=3},
  ipe mark large/.style={ipe mark scale=5},
  ipe mark small/.style={ipe mark scale=2},
  ipe mark tiny/.style={ipe mark scale=1.1},
  ipe mark normal,
  /pgf/arrow keys/.cd,
  ipe arrow normal/.style={scale=7},
  ipe arrow large/.style={scale=10},
  ipe arrow small/.style={scale=5},
  ipe arrow tiny/.style={scale=3},
  ipe arrow normal,
  /tikz/.cd,
  ipe arrows, 
  <->/.tip = ipe normal,
  ipe dash normal/.style={dash pattern=},
  ipe dash dashed/.style={dash pattern=on 4bp off 4bp},
  ipe dash dotted/.style={dash pattern=on 1bp off 3bp},
  ipe dash dash dotted/.style={dash pattern=on 4bp off 2bp on 1bp off 2bp},
  ipe dash dash dot dotted/.style={dash pattern=on 4bp off 2bp on 1bp off 2bp on 1bp off 2bp},
  ipe dash normal,
  ipe node/.append style={font=\normalsize},
  ipe stretch normal/.style={ipe node stretch=1},
  ipe stretch normal,
  ipe opacity 10/.style={opacity=0.1},
  ipe opacity 30/.style={opacity=0.3},
  ipe opacity 50/.style={opacity=0.5},
  ipe opacity 75/.style={opacity=0.75},
  ipe opacity opaque/.style={opacity=1},
  ipe opacity opaque,
]
\definecolor{red}{rgb}{1,0,0}
\definecolor{green}{rgb}{0,1,0}
\definecolor{blue}{rgb}{0,0,1}
\definecolor{yellow}{rgb}{1,1,0}
\definecolor{orange}{rgb}{1,0.647,0}
\definecolor{gold}{rgb}{1,0.843,0}
\definecolor{purple}{rgb}{0.627,0.125,0.941}
\definecolor{gray}{rgb}{0.745,0.745,0.745}
\definecolor{brown}{rgb}{0.647,0.165,0.165}
\definecolor{navy}{rgb}{0,0,0.502}
\definecolor{pink}{rgb}{1,0.753,0.796}
\definecolor{seagreen}{rgb}{0.18,0.545,0.341}
\definecolor{turquoise}{rgb}{0.251,0.878,0.816}
\definecolor{violet}{rgb}{0.933,0.51,0.933}
\definecolor{darkblue}{rgb}{0,0,0.545}
\definecolor{darkcyan}{rgb}{0,0.545,0.545}
\definecolor{darkgray}{rgb}{0.663,0.663,0.663}
\definecolor{darkgreen}{rgb}{0,0.392,0}
\definecolor{darkmagenta}{rgb}{0.545,0,0.545}
\definecolor{darkorange}{rgb}{1,0.549,0}
\definecolor{darkred}{rgb}{0.545,0,0}
\definecolor{lightblue}{rgb}{0.678,0.847,0.902}
\definecolor{lightcyan}{rgb}{0.878,1,1}
\definecolor{lightgray}{rgb}{0.827,0.827,0.827}
\definecolor{lightgreen}{rgb}{0.565,0.933,0.565}
\definecolor{lightyellow}{rgb}{1,1,0.878}
\definecolor{black}{rgb}{0,0,0}
\definecolor{white}{rgb}{1,1,1}
\begin{tikzpicture}[ipe stylesheet]
  \filldraw[shift={(16, 845.025)}, xscale=0.8715, yscale=1.4535, black, ipe pen heavier]
    (0, 0) rectangle (384, -64);
  \filldraw[shift={(266.984, 845.025)}, xscale=0.8715, yscale=1.4535, ipe pen heavier, fill=yellow]
    (0, 0) rectangle (64, 16);
  \filldraw[shift={(43.887, 845.025)}, xscale=0.8715, yscale=1.4535, ipe pen heavier, fill=yellow]
    (0, 0) rectangle (64, 16);
  \draw[shift={(16, 845.024)}, xscale=0.8715, yscale=1.2409, ipe pen heavier]
    (0, 0) rectangle (384, 64);
  \pic[ipe mark large, darkgreen]
     at (30.3698, 965.1171) {ipe disk};
  \pic[ipe mark large, darkgreen]
     at (46.4931, 950.4491) {ipe disk};
  \pic[ipe mark large, darkgreen]
     at (61.8835, 970.6171) {ipe disk};
  \pic[ipe mark large, darkgreen]
     at (75.0752, 948.6151) {ipe disk};
  \pic[ipe mark large, darkgreen]
     at (95.2292, 970.6171) {ipe disk};
  \pic[ipe mark large, darkgreen]
     at (108.0541, 949.8381) {ipe disk};
  \pic[ipe mark large, darkgreen]
     at (123.4451, 963.8941) {ipe disk};
  \pic[ipe mark large, darkgreen]
     at (135.5371, 954.7271) {ipe disk};
  \pic[ipe mark large, darkgreen]
     at (152.7601, 965.1171) {ipe disk};
  \pic[ipe mark large, darkgreen]
     at (153.8591, 946.7821) {ipe disk};
  \pic[ipe mark large, darkgreen]
     at (174.7461, 973.0621) {ipe disk};
  \pic[ipe mark large, darkgreen]
     at (180.6091, 944.9481) {ipe disk};
  \pic[ipe mark large, darkgreen]
     at (198.5641, 970.6171) {ipe disk};
  \pic[ipe mark large, darkgreen]
     at (206.2591, 949.2261) {ipe disk};
  \pic[ipe mark large, darkgreen]
     at (224.9481, 968.7841) {ipe disk};
  \pic[ipe mark large, darkgreen]
     at (220.5501, 943.7261) {ipe disk};
  \pic[ipe mark large, darkgreen]
     at (257.5601, 960.8391) {ipe disk};
  \pic[ipe mark large, darkgreen]
     at (238.8721, 948.0041) {ipe disk};
  \pic[ipe mark large, darkgreen]
     at (243.2691, 973.6731) {ipe disk};
  \pic[ipe mark large, darkgreen]
     at (272.2181, 978.5631) {ipe disk};
  \pic[ipe mark large, darkgreen]
     at (261.5911, 947.3931) {ipe disk};
  \pic[ipe mark large, darkgreen]
     at (278.4471, 954.1161) {ipe disk};
  \pic[ipe mark large, darkgreen]
     at (295.3031, 977.3401) {ipe disk};
  \pic[ipe mark large, darkgreen]
     at (292.0051, 956.5601) {ipe disk};
  \pic[ipe mark large, darkgreen]
     at (304.4641, 943.7261) {ipe disk};
  \pic[ipe mark large, darkgreen]
     at (313.6251, 973.6731) {ipe disk};
  \pic[ipe mark large, darkgreen]
     at (316.5571, 957.1721) {ipe disk};
  \pic[ipe mark large, darkgreen]
     at (329.3821, 948.6151) {ipe disk};
  \pic[ipe mark large, darkgreen]
     at (330.4811, 973.6731) {ipe disk};
  \pic[ipe mark large, darkgreen]
     at (342.2071, 948.0041) {ipe disk};
  \pic[ipe mark large, darkgreen]
     at (344.4061, 973.0621) {ipe disk};
  \pic[ipe mark large, darkgreen]
     at (210.2901, 977.3401) {ipe disk};
  \pic[ipe mark large, darkgreen]
     at (168.8831, 958.3941) {ipe disk};
  \pic[ipe mark large, darkgreen]
     at (106.2221, 977.9511) {ipe disk};
  \pic[ipe mark large, darkgreen]
     at (134.8041, 976.1181) {ipe disk};
  \pic[ipe mark large, darkgreen]
     at (91.5648, 948.0041) {ipe disk};
  \pic[ipe mark large, darkgreen]
     at (79.106, 973.6731) {ipe disk};
  \pic[ipe mark large, darkgreen]
     at (62.9828, 948.6151) {ipe disk};
  \pic[ipe mark large, darkgreen]
     at (45.7603, 974.8961) {ipe disk};
  \pic[ipe mark large, darkgreen]
     at (23.0412, 948.6151) {ipe disk};
  \pic[ipe mark large, darkgreen]
     at (21.9419, 976.1181) {ipe disk};
  \filldraw[shift={(16, 1007.818)}, xscale=0.8715, yscale=1.4535, black]
    (0, 0) rectangle (384, -16);
  \draw[shift={(16.328, 984.561)}, xscale=0.8706, yscale=1.8775, ipe pen heavier]
    (0, 0) rectangle (384, -32);
  \draw[ipe pen heavier]
    (332.4694, 825.9261)
     -- (444.4694, 825.9261);
  \draw[ipe pen heavier]
    (305.1305, 860.8731)
     -- (461.4655, 860.3271);
  \draw[ipe pen heavier]
    (332.4694, 900.5853)
     -- (396.4694, 900.5853);
  \draw[shift={(334.583, 933.672)}, xscale=1.0774, yscale=-0.3693, ipe pen heavier]
    (0, 0)
     -- (137.838, -0.249);
  \draw[ipe pen heavier]
    (341.4092, 963.6887)
     -- (393.9692, 963.6887);
  \draw[ipe pen heavier]
    (328.0678, 994.4722)
     -- (388.1958, 994.4722);
  \node[ipe node, font=\huge]
     at (447.182, 820.944) {\textbf{Fisheye Camera}};
  \node[ipe node, font=\huge]
     at (464.215, 856.244) {\textbf{LEDs}};
  \node[ipe node, font=\huge]
     at (399.288, 898.838) {\textbf{Stiff silicone layer (17 mm)}};
  \node[ipe node, font=\huge]
     at (485.717, 926.263) {\textbf{Markers}};
  \node[ipe node, font=\huge]
     at (395.882, 960.846) {\textbf{Transparent gel (4.5 mm)}};
  \node[ipe node, font=\huge]
     at (390.677, 991.529) {\textbf{Black silicone layer (1.5 mm)}};
  \filldraw[black, ipe pen heavier]
    (134.2911, 844.0629)
     arc[start angle=-173.6598, end angle=-6.3402, x radius=50.7101, y radius=-52.3198];
  \pic[ipe mark large, darkgreen]
     at (33.3971, 934.4141) {ipe disk};
  \pic[ipe mark large, darkgreen]
     at (57.0657, 933.2311) {ipe disk};
  \pic[ipe mark large, darkgreen]
     at (81.9178, 936.1891) {ipe disk};
  \pic[ipe mark large, darkgreen]
     at (109.1371, 933.8221) {ipe disk};
  \pic[ipe mark large, darkgreen]
     at (133.3971, 932.0471) {ipe disk};
  \pic[ipe mark large, darkgreen]
     at (122.1541, 945.6571) {ipe disk};
  \pic[ipe mark large, darkgreen]
     at (165.9411, 933.8221) {ipe disk};
  \pic[ipe mark large, darkgreen]
     at (204.4031, 963.4081) {ipe disk};
  \pic[ipe mark large, darkgreen]
     at (187.2431, 957.4911) {ipe disk};
  \pic[ipe mark large, darkgreen]
     at (197.8941, 930.8641) {ipe disk};
  \pic[ipe mark large, darkgreen]
     at (219.1961, 928.4971) {ipe disk};
  \pic[ipe mark large, darkgreen]
     at (231.0301, 936.1891) {ipe disk};
  \pic[ipe mark large, darkgreen]
     at (258.8411, 931.4561) {ipe disk};
  \pic[ipe mark large, darkgreen]
     at (272.4501, 933.2311) {ipe disk};
  \pic[ipe mark large, darkgreen]
     at (290.2021, 930.8641) {ipe disk};
  \pic[ipe mark large, darkgreen]
     at (324.5211, 942.1061) {ipe disk};
  \pic[ipe mark large, darkgreen]
     at (335.1721, 932.0471) {ipe disk};
  \pic[ipe mark large, darkgreen]
     at (306.7701, 933.2311) {ipe disk};
\end{tikzpicture}}
	\caption{The full thickness of the experimental setup from the base of the camera to the surface is approximately 38 mm. The thickness of the different silicone layers is indicated in the figure.}
	\label{fig:sensor_scheme}
\end{figure}

The tactile sensor discussed here, schematically shown in Fig. \ref{fig:sensor_scheme}, is made of a silicone gel (with a squared horizontal section of 32x32 mm) with spherical markers spread over its volume and a camera underneath to track the motion of these markers.
The camera is an ELP USBFHD06H, equipped with a fisheye lens that has an angle of view of 180 degrees. The image frames are acquired at a resolution of 640x480 pixels and cropped to a region of interest of 440x440 pixels. The camera frame rate is 60 fps. 

The soft silicone gel is produced in a three-layer structure as described in \cite{sferrazza_sensors}, with the difference that smaller fluorescent green polyethylene microspheres (with a diameter of 150 to 180 $\mu$m) are used as markers to provide strain information at a finer spatial resolution. An indication of the thickness of the different layers is shown in Fig. \ref{fig:sensor_scheme}.

The randomness of the marker positions over the gel's volume facilitates production. Conversely to sensing techniques that require a specific pattern in the marker distribution, the sensor discussed here is directly adaptable to any required shape.

Note that the design proposed shows a proof of concept and its dimensions have not been optimized. Nevertheless, the size is suitable for deployment to a robot gripper, as shown in \cite{gelsight_gripper} with a vision-based sensor of similar dimensions. The current sensor thickness is mainly determined by the fisheye lens and the stiff layer. The thickness of these components can be traded off against the field of view of the camera, i.e., smaller cameras with lower angles of view and placed closer to the sensor's surface would result in reduced thickness but smaller gel coverage. In this respect, the use of a higher number of cameras can increase the field of view while retaining a limited thickness. Alternatively, applications with very thin structures, e.g. robotic fingers, can be addressed by the use of mirrors \cite{gelslim}.
In the context of robot skins, a coarser spatial resolution is generally required in sections of the body where space is of less relevance, i.e. humanoid robot trunk or arms, where tactile sensors are generally used for collision detection. As an example, the camera might be placed inside the robot trunk at a greater distance from the surface to cover a larger portion of the body.

\section{DATA COLLECTION} \label{sec:data_collection}
\begin{figure}
	\includegraphics[width=\linewidth]{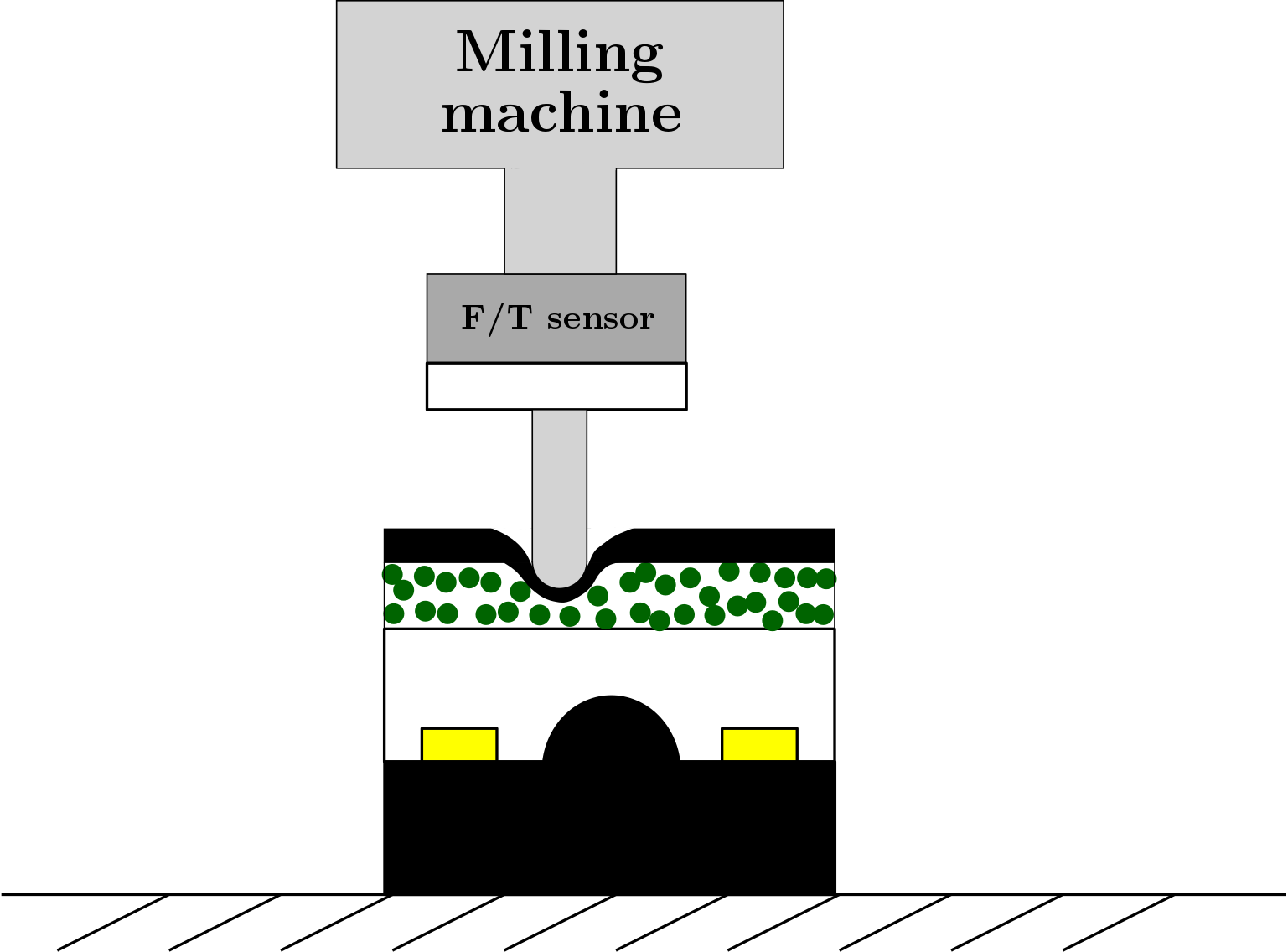}
	\caption{The indenter is controlled by a CNC milling machine and it is pressed against the gel while the data from the camera and the F/T sensor are recorded.}
	\label{fig:data_collection}
\end{figure}

The neural network architecture needs a considerable amount of training data to be able to predict the normal force distribution with satisfactory accuracy. In order to automatize the data collection procedure, a precision milling and drilling machine (Fehlmann PICOMAX 56 TOP) with 3-axis computer numerical control (CNC) is used. A F/T sensor (ATI Mini40) is attached to the spindle of the machine to assign ground truth force labels to the data. A plexiglass plate, connected to the F/T sensor, serves as a holder for a spherical-ended cylindrical indenter that is approximately 40 mm long with a diameter of 1.2 mm. The tactile sensor is clamped to the machine base, and the indenter is pressed and released over the entire surface of the gel at different depths to span a certain range of normal force values. The F/T readings are recorded while the camera streams the pictures at a fixed frequency. The milling machine provides a 24 V digital signal that is used to synchronize the data from the different sources. Fig. \ref{fig:data_collection} shows the data collection setup.

\section{FEATURE ENGINEERING} \label{sec:feature_engineering}
The task of reconstructing the normal force distribution on the surface of the tactile sensor can be formulated as a supervised learning problem, mapping the information obtained from the camera images to the applied force. To this purpose, the selection of relevant features is of great importance. Similarly, choosing an appropriate representation of the force distribution, that is, the labels of the supervised learning algorithm, is crucial.

In order to have a flexible representation, suitable to span the force fields generated by multiple objects with arbitrary shapes, the approach presented in \cite{sferrazza_sensors} and used in this paper consists of the discretization of the gel's surface in $n$ smaller bins, assigning a force value to each of these. The resulting $n$ values for each data point are stacked together in a $n$-dimensional vector, which represents the label in that instance. 

\begin{figure}[h!]
	\centering
	\includegraphics[width=0.4\linewidth]{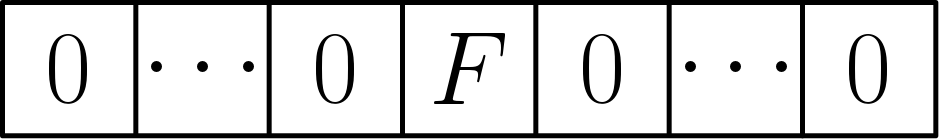}
	\caption{For the spherical indenter described in Section \ref{sec:data_collection}, the resulting normal force distribution is nonzero only at the bin that includes the indentation position, where it takes the value measured by the F/T sensor, here indicated with $F$.}
	\label{fig:label_vector}
\end{figure}

In the special case of the data collected with the automatic procedure described in Section \ref{sec:data_collection}, this vector has a very sparse structure. In fact, assuming that the surface in contact with the indenter lies entirely inside one bin, the value of the normal force applied by the tip of the needle on the gel's surface is encoded at the vector's component representing the bin that contains the point of application. The remaining vector components are filled with zeros. An example of a label vector obtained from the automatic data procedure is shown in Fig. \ref{fig:label_vector}. Note that the previous assumption might be violated for indentation cases at the boundaries of the bins. In these cases, the representation above approximates the normal force distribution by assigning the entire measured force to the bin containing the indentation center.

With regard to the input to the force reconstruction model, the images are processed before being fed to the learning architecture for training or prediction. Conversely to directly taking the image pixels as features, extracting meaningful information from each image and creating a more compact set of features results in lower data requirements and shorter training times. In addition, having features that are likely to be invariant under the same force distribution, when extracted on separate gels with different marker patterns, is highly desirable. In this respect, this paper discusses an approach based on dense optical flow \cite{optical_flow_book}.

Dense optical flow algorithms aim to estimate the motion at each pixel of the image. The algorithm based on Dense Inverse Search (DIS), see \cite{DIS}, approaches this task by reconstructing the flow at multiple scales in a coarse-to-fine fashion. DIS employs an efficient search of correspondences, which renders the approach suitable to run in real-time. With respect to the application proposed here, the algorithm is independent of the marker type and distribution, since it only requires a trackable distinct pattern, which in the tactile sensor discussed in this paper is in fact given by the spherical markers. 

\begin{figure}
	\centering
	\subfloat[Original image]{%
		\includegraphics[width=0.48\linewidth]{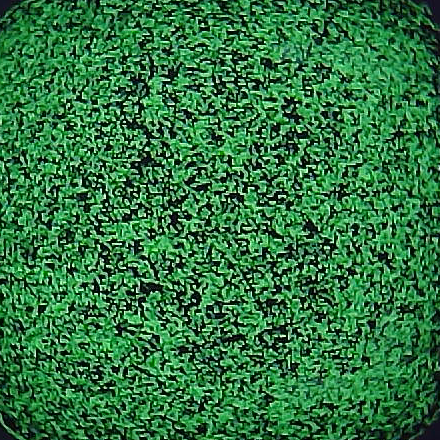} }
	\hfill
	\subfloat[Optical flow]{%
		\includegraphics[width=0.48\linewidth]{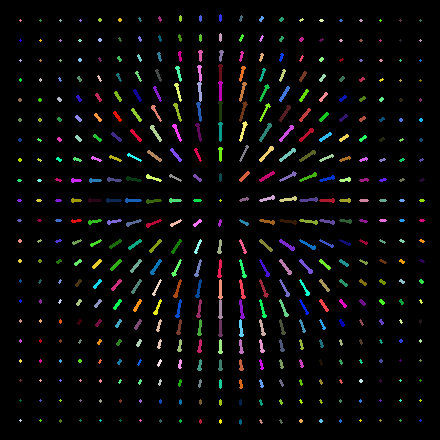} }
	\caption{The RGB camera image, captured at rest in (a), is first converted to grayscale. The dense optical flow, shown in (b) at subsampled locations for an example indentation, is then computed with respect to the image at rest using the DIS-based algorithm.}
	\label{fig:dense_optical_flow}
\end{figure}

The original RGB image is converted to grayscale, and the optical flow algorithm is applied. An example of the computed flow is shown in Fig. \ref{fig:dense_optical_flow}.

The resulting flow is stored at each pixel as tuples of magnitude and an angle with respect to a fixed axis. As in \cite{sferrazza_sensors}, the image plane is then divided into a grid of $m$ regions with equal area and each optical flow tuple is assigned to one of these regions, depending on its position in the image. For each region $i = 1, \dots, m$, the average magnitude and direction of the optical flow are computed. These average quantities have the advantage of implicitly storing position information (in fact, they are an approximation of the optical flow at each of the regions' centers). Furthermore, provided that a distinct pattern is available at each region, they are independent of the markers' distribution over the entire volume. Therefore, they are designed to be invariant when the same forces are applied to a different gel with the same material properties but not necessarily with the same marker pattern.
The two average quantities for each of the $m$ regions are chosen as the $2\times m$ features, which are the input to the neural network architecture that is presented in Section \ref{sec:neural_network}.
\section{MODEL TRAINING} \label{sec:neural_network}
\subsection{Learning architecture}
\begin{figure}
	\includegraphics[width=\linewidth]{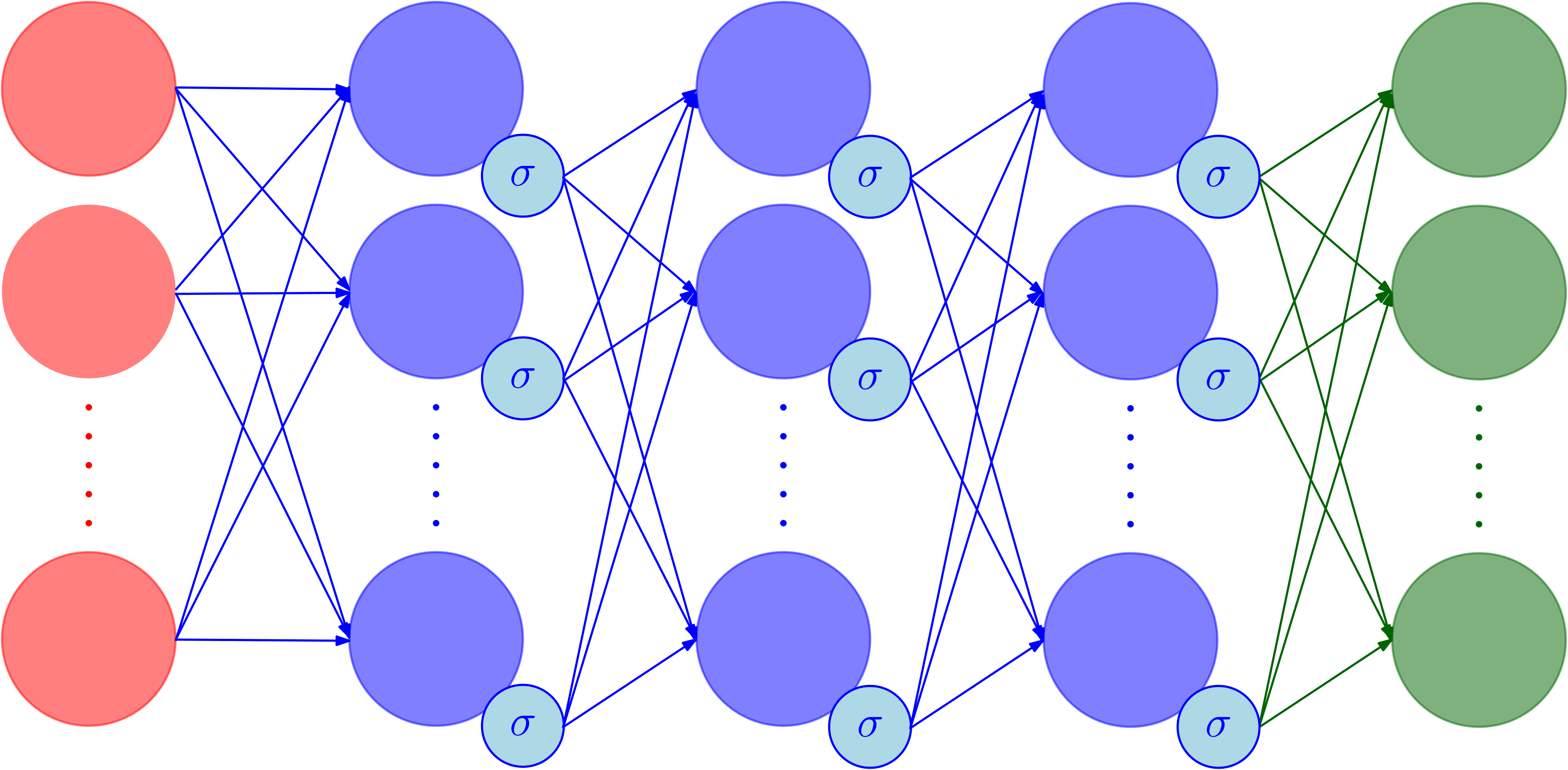}
	\caption{The DNN architecture. In red the input layer (with $2\times m$ placeholders), in blue the hidden layers with the logistic activation functions, indicated with $\sigma$, in green the output layer (with $n$ neurons), which has an identity activation function to perform the regression task. The biases are not shown in the figure.}
	\label{fig:original_architecture}
\end{figure}

A deep neural network (DNN) is designed to estimate the normal force distribution applied to the tactile sensor's surface, given the features extracted from the optical flow. 
A feedforward architecture with three fully connected hidden layers, all using a logistic function as the activation unit, is chosen. The input layer reads the $2\times m$ features described in Section \ref{sec:feature_engineering}, while the output layer returns the predicted $n$-dimensional output, which assigns a force value to each of the sensor's surface bins. A scheme summarizing the chosen architecture, which exhibits relatively fast training times, is shown in Fig. \ref{fig:original_architecture}.

The loss used to train the DNN is the average root mean squared error ($aRMSE$, see \cite{multioutput_survey}), that is,
\begin{equation}
aRMSE = \frac{1}{n} \sum_{i=0}^{n\minus1}\sqrt{\frac{\sum_{l=0}^{N_{\text{set}}\minus1}\left(y_i^{(l)}-\hat{y}_i^{(l)}\right)^2}{N_{\text{set}}}},
\end{equation}
where $y_i^{(l)}$ and $\hat{y}_i^{(l)}$ are the $i$-th components of the true and the predicted $l$-th label vector, respectively, and $N_\text{set}$ is the number of samples in the set that is being evaluated (i.e. training, validation, test sets).
The optimization method used to train the network is RMSProp with Nesterov momentum (known as Nadam, see \cite{nadam}).
In order to prevent overfitting, dropout layers, see \cite{dropout}, are added after each of the hidden layers and are used during the training phase. Moreover, a portion of the training data is selected as a validation set, and the loss computed on this set is used to early stop the optimization when this loss does not decrease for $N_{\text{es}}$ consecutive epochs. 

\subsection{Evaluation}

A dataset is collected using the automatic procedure presented in Section \ref{sec:data_collection}. The needle is pressed on the square surface at a set of positions described by a grid with equal spacings of 0.75 mm. The tip of the indenter reaches eight different depth levels, from 0.25 mm to 2 mm. This procedure gives 10952 data points, with the normal force measured by the F/T sensor up to 1 N.

The other parameters used in this experiment are summarized in Table \ref{table:parameters}.

\begin{table}[h!]
	\centering
	\begin{tabular}{c|c|c} 
		\hline
		\rule{0pt}{2ex}
		Symbol & Value & Description \\
		\hline
		\rule{0pt}{3ex}
		$m$ & 1600 & \# of averaging regions in the image\\ 
		$n$ & 81 & \# of bins that grid the sensor surface\\
		- & 200 & training batch size\\
		- & 0.001 & learning rate\\
		$N_{\text{es}}$ & 50 & early stopping parameter \\
		- & 0.15 & dropout rate \\
		- & (800,400,400) & hidden layers' size\\ [1ex] 
		\hline
	\end{tabular}
	\caption{Parameters used to train the DNN architecture.}
	\label{table:parameters}
\end{table}

Before training, 20\% of the dataset is put aside as a test set, that is subsequently used to evaluate the results. 

Given the sparse nature of the label vectors, the evaluation of the results is discussed with respect to some quantities (additionally to the $aRMSE$) that are particularly intuitive for this application. These measures have been introduced in \cite{sferrazza_sensors} and capture the performance of the sensor in predicting the correct location and magnitude of the force applied with the needle used for this experiment. For $l = 0, 1, \dots, N_\text{set}-1$, the following indexes are computed,
\begin{align} \notag
k_l = \argmax_{i=0,1,\dots,n-1} |y_i^{(l)}|, \qquad 
\hat{k}_l = \argmax_{i=0,1,\dots,n-1} |\hat{y}_i^{(l)}|.
\end{align}
An error metric based on the distance between the maximum components of the true and estimated label vectors is then introduced as,
\begin{align}
d_{\text{loc}} = \frac{1}{N_\text{set}} \sum_{l=0}^{N_\text{set}-1}\|c(k_l) - c(\hat{k}_l)\|_2,
\label{eq:distance_metric}
\end{align}
where $c(k)$ denotes the location of the center of the bin $k$ on the surface.
Similarly, an error that only considers the magnitude of the maximum components of the true and estimated label vectors is defined as,
\begin{align}
RMSE_{\text{mc}}  = \sqrt{\frac{1}{N_\text{set}} \sum_{l=0}^{N_\text{set}-1} \left(y_{k_l}^{(l)}-\hat{y}_{\hat{k}_l}^{(l)}\right)^2}.
\label{eq:max_metric}
\end{align}

\begin{figure}
	\centering
	\includegraphics[width=\linewidth]{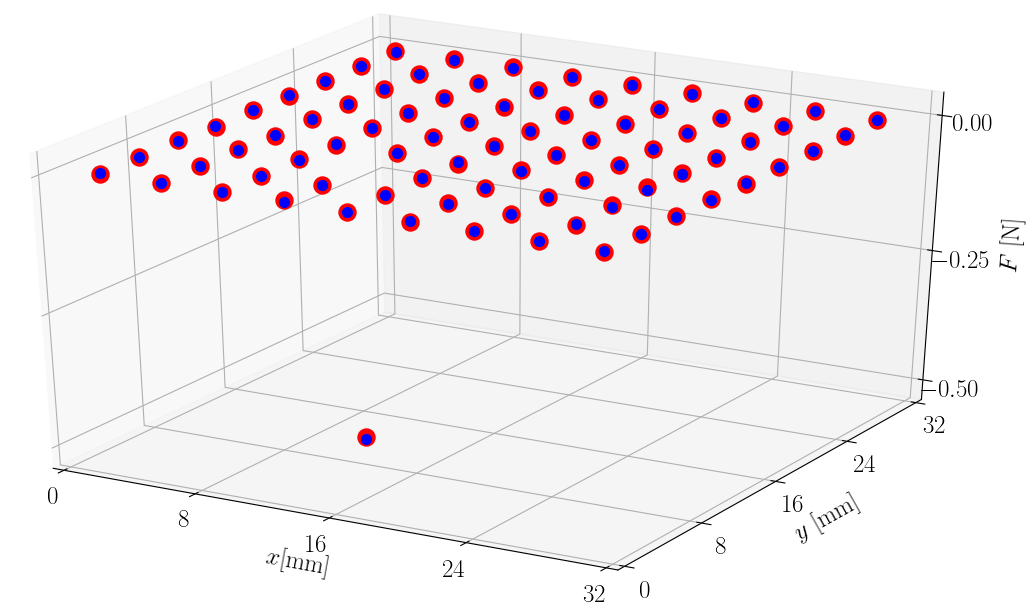}
	\caption{The predicted normal force distribution (in red) for a sample in the test set, compared to the ground truth (in blue). $x$ and $y$ are the two Cartesian axes spanning the gel's surface, with the origin at one corner. The sign of the normal force $F$ is defined to be negative when directed towards the camera.}
	\label{fig:predicted_sample}
\end{figure}

The results of the trained model evaluated on the test set are summarized in Table \ref{table:dnn_results}. An example of the predicted normal force distribution is shown in Fig. \ref{fig:predicted_sample}.

\begin{table}[h!]
	\centering
	\begin{tabular}{c|c|c} 
		\hline
		\rule{0pt}{2ex}
		Criterion & Value & Unit \\
		\hline
		\rule{0pt}{3ex}
		$aRMSE$ & 0.009 & [N]\\ 
		$d_\text{loc}$ & 0.107 & [mm]\\
		$RMSE_\text{mc}$ & 0.065 & [N]\\
		[1ex] 
		\hline
	\end{tabular}
	\caption{Evaluation of the trained model on the test set.}
	\label{table:dnn_results}
\end{table}

\section{CALIBRATION} \label{sec:calibration}
Collecting large datasets for each sensor is time consuming, even if automatized as described in Section \ref{sec:data_collection}. Furthermore, a model that has been trained to reconstruct the normal force distribution on a particular gel may not yield the same prediction performance on another gel. Despite the sensor design and the invariance of the chosen features, there are still other variables that are specific for each sensor and might not have been accounted for.

Modeling the gel as a linearly elastic half-space, the magnitude of the 3D elastic displacement $u$ of a marker is related to the concentrated normal force $F$ applied to the sensor's surface by a proportional factor, that is,
\begin{equation}
u = h(E) F,
\end{equation}
where $h$ depends on the hardness of the material, through its Young's modulus $E$. A formal derivation of this fact is given by the Bousinnesq solution \cite[p.~50]{contact_mechanics}. However, the resulting hardness of the gel's surface is sensitive to the actual percentage of the two components that are mixed together to produce it. When the model trained on one gel is applied to a gel with a different hardness (i.e. mixing ratio), an appropriate transformation of the marker displacements observed in the images is necessary to attain comparable performance. Nevertheless, it is not straightforward to perform this operation due to the influence of the camera model on this transformation and to the presence of noise introduced by the camera acquisition and the optical flow computation.

In addition, the relative position of the gel with respect to the camera is crucial. In fact, the sensitivity of a marker's position in the image (with respect to a fixed image coordinate frame) to changes in the distance $z$ from the lens is inversely proportional to the square of $z$. That is, from the pinhole camera equations (see \cite[p.~49]{computervision_book}),
\begin{equation}
p \propto \frac{1}{z} \Rightarrow \partder{p}{z} \propto \frac{1}{z^2},
\end{equation}
where $p$ is the distance of the marker in the image from the origin of the image coordinate frame. 
Therefore, given the small distance of the markers from the fisheye lens, relatively small differences in the thickness of the materials across multiple gels (that are introduced during the production and assembly of the sensor) result in considerably different observed displacements.

Moreover, if a different camera is used, its intrinsic parameters have an effect on how the marker displacements are projected onto the image plane. This problem can be solved by undistorting the images, a procedure that may however introduce other inaccuracies in compensating for the considerable distortion of the lens.

Considering that the differences mentioned are all at the level of the features, a preprocessing fully connected layer with a rectified linear unit (ReLU) activation function is added to the previously trained DNN architecture between the input and the first hidden layer. The weight matrix corresponding to this preprocessing layer is initialized with the identity, which leads to a considerable speed up in learning. In fact, the differences across sensors are expected to contribute to rather small deviations from a diagonal structure of the weight matrix.

\begin{figure}
	\centering
	\includegraphics[width=0.6\linewidth]{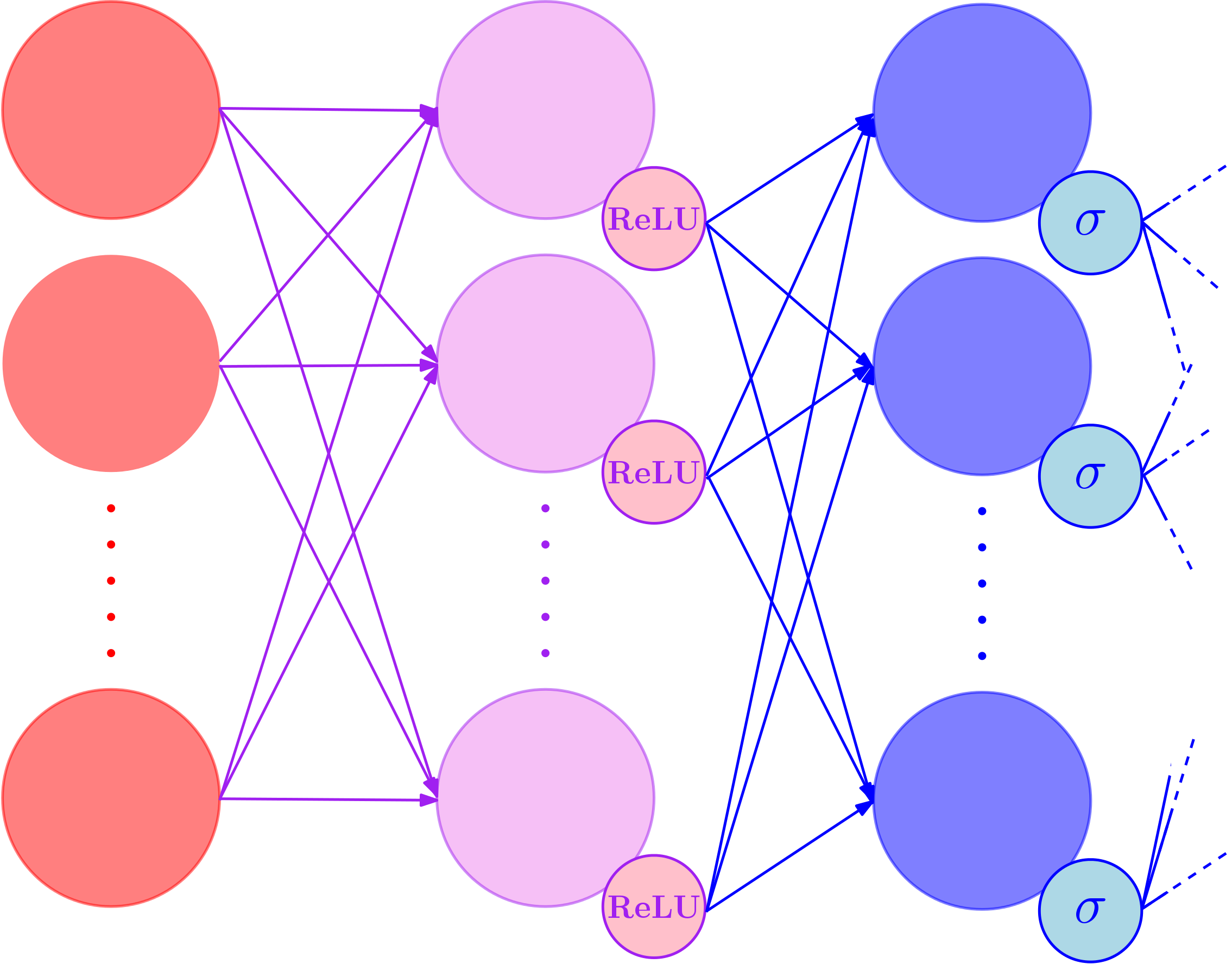}
	\caption{The input layer is directed towards a calibration layer (in violet) with $2\times m$ neurons, which is then connected to the hidden layers of the original architecture.}
	\label{fig:calibration_architecture}
\end{figure}

\begin{figure}
	\hspace{-0.5cm}
	\includegraphics[width=1.1\linewidth]{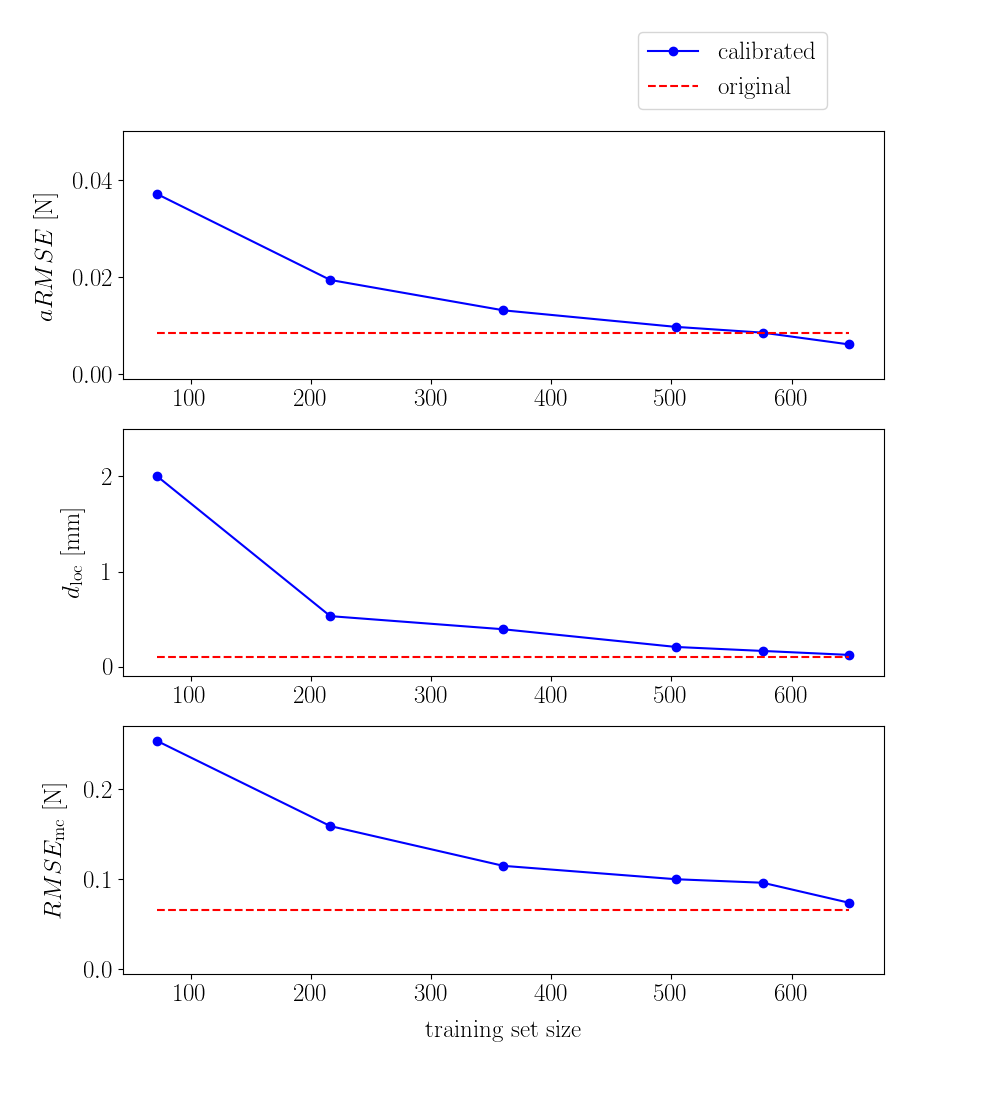}
	\vspace{-1cm}
	\caption{A dataset of 800 samples is collected on the sensor that undergoes the calibration procedure. Portions of different sizes are used as training sets, while the remaining data serve as a test set for evaluation. The different metrics show that the augmented architecture (in blue) attains comparable performance to the experiment presented in Section \ref{sec:neural_network} (in red), where the network was trained with a much larger dataset (7884 data points). Training both the original and the augmented architecture with the smaller dataset from scratch yields a substantially inferior performance, and the results are therefore not shown in this plot. The same applies to predicting the normal force distribution on the new gel with the network trained in Section \ref{sec:neural_network} (on the first gel, before calibration).}
	\label{fig:calibration_results}
\end{figure}

A considerably smaller number of data points is collected on a new gel (over a coarser grid), and the augmented architecture (of which a snippet is shown in Fig. \ref{fig:calibration_architecture}) is then trained on this dataset by freezing all the weights apart from the ones belonging to the added preprocessing layer. In this way, the training time and the data requirements are greatly reduced, while retaining comparable performance to the one obtained in Section \ref{sec:neural_network}. The parameters used for this procedure are summarized in Table \ref{table:calibration_parameters} and the results are shown in Fig. \ref{fig:calibration_results}, for different sizes of the training set.

\begin{table}[h!]
	\centering
	\begin{tabular}{c|c|c} 
		\hline
		\rule{0pt}{2ex}
		Symbol & Value & Description \\
		\hline
		\rule{0pt}{3ex}
		- & 64 & size of training batches\\
		- & 0.0001 & learning rate\\
		$N_{\text{es}}$ & 200 & early stopping parameter \\
		- & 0.05 & dropout rate \\ 
		[1ex] 
		\hline
	\end{tabular}
	\caption{Parameters used to train the calibration layer}
	\label{table:calibration_parameters}
\end{table}

Note that the success of this rather simple calibration technique is mainly due to the choice of the features described in Section \ref{sec:feature_engineering}. As opposed to learning the force reconstruction task directly from the pixel values, the averaged optical flow features are in fact invariant across different gels, except for the alignment and scaling factors mentioned above.

The algorithms presented in this paper yield a real-time execution of 60 Hz, with the implementation not particularly optimized for efficiency. An overview of the time needed for a single execution of the main steps of the pipeline is shown in Table \ref{table:prediction_times}. 

\begin{table}[h!]
	\centering
	\begin{tabular}{c|c} 
		\hline
		\rule{0pt}{2ex}
		Component & Time (ms) \\
		\hline
		\rule{0pt}{3ex}
		Image acquisition and cropping & 1 \\
		Optical flow computation & 9\\
		Feature generation & 2\\
		Prediction & 2 \\
		[1ex] 
		\hline
	\end{tabular}
	\caption{Average times for real-time pipeline}
	\label{table:prediction_times}
\end{table}

\section{CONCLUSION} \label{sec:conclusion}
An approach to sense the normal force distribution on a soft surface has been discussed. A learning algorithm has been applied to a vision-based tactile sensor, which is inexpensive and easy to manufacture. The proposed strategy is scalable to arbitrary surfaces and therefore suitable for robot skin applications. 

The results (see the video accompanying this paper) show that the sensor can reconstruct the normal force distribution applied with a test indenter after being trained on an automatically collected dataset. Note that the learning problem discussed here is a multiple multivariate regression, which maps multi-dimensional feature vectors to multi-dimensional label vectors. The DNN architecture is able to predict the force applied to each of the surface bins, which can either be zero or the one actually applied with the point indenter. Conversely to regression techniques that separately predict each output, a feedforward neural network can capture the interrelations between the different output label components. Since the network does not directly use knowledge of the point indentation, this strategy is therefore appealing for more general contacts (e.g. with larger indenters or multiple contacts) that will be the subject of future work.

The tactile sensing pipeline proposed here estimates the static force distribution from the images captured by a camera. The training datasets employed in this paper capture measurements taken after the indentation has reached steady-state. Nevertheless, dynamic indentations are reconstructed with a limited loss of accuracy, as shown in the accompanying video. As an example, a detailed analysis of hysteresis effects has the potential of further improving the predictions in both the loading and unloading phases.

In order to speed up the learning and the training data collection, the variations across different sensors of the same type have been identified at the level of the features. Therefore, this paper has proposed a calibration procedure that accordingly modifies the input to the learning architecture, thus transferring the knowledge across different gels. 

\addtolength{\textheight}{0cm}   





\section*{ACKNOWLEDGMENT}

The authors would like to thank Michael Egli and Marc-Andr\'e Corzillius for their support in the manufacturing of the sensor, and Laura Gasser for her contribution to the feature engineering.


\bibliographystyle{IEEEtran}
\bibliography{IEEEabrv,references}

\end{document}